\documentclass[runningheads]{llncs}
\usepackage{graphicx}
\usepackage{amsmath}
\usepackage{amssymb}
\usepackage{float}
\usepackage{lscape}

\DeclareMathOperator*{\argmin}{arg\,min}
\newcommand{\norm}[1]{\left\lVert#1\right\rVert}
\usepackage{algorithm2e}
\RestyleAlgo{ruled} 
\usepackage{algpseudocode}
\usepackage{caption}
\usepackage{subcaption}
\usepackage{xcolor}
\usepackage{multirow}
\usepackage{hyperref}
\newcommand\crule[3][black]{\textcolor{#1}{\rule{#2}{#3}}}
\begin{document}
\title{Driving Points Prediction\\ For Abdominal Probabilistic Registration}
\titlerunning{Driving Points Prediction For Abdominal Probabilistic Registration}
%

\author{Samuel Joutard, Reuben Dorent, Sebastien Ourselin, \\Tom Vercauteren, Marc Modat}

\authorrunning{S. Joutard}
%
\institute{King's College London}
\maketitle              
\begin{abstract}
Inter-patient abdominal registration has various applications, from pharmakinematic studies to anatomy modeling. Yet, it remains a challenging application due to the morphological heterogeneity and variability of the human abdomen. Among the various registration methods proposed for this task, probabilistic displacement registration models estimate displacement distribution for a subset of points by comparing feature vectors of points from the two images. These probabilistic models are informative and robust while allowing large displacements by design. As the displacement distributions are typically estimated on a subset of points (which we refer to as driving points), due to computational requirements, we propose in this work to learn a driving points predictor. Compared to previously proposed methods, the driving points predictor is optimized in an end-to-end fashion to infer driving points tailored for a specific registration pipeline. We evaluate the impact of our contribution on two different datasets corresponding to different modalities. Specifically, we compared the performances of 6 different probabilistic displacement registration models when using a driving points predictor or one of 2 other standard driving points selection methods. The proposed method improved performances in 11 out of 12 experiments. 
\end{abstract}
%


\section{Introduction}
\label{sec:intro}
\begin{figure}[t]
\includegraphics[width=\linewidth]{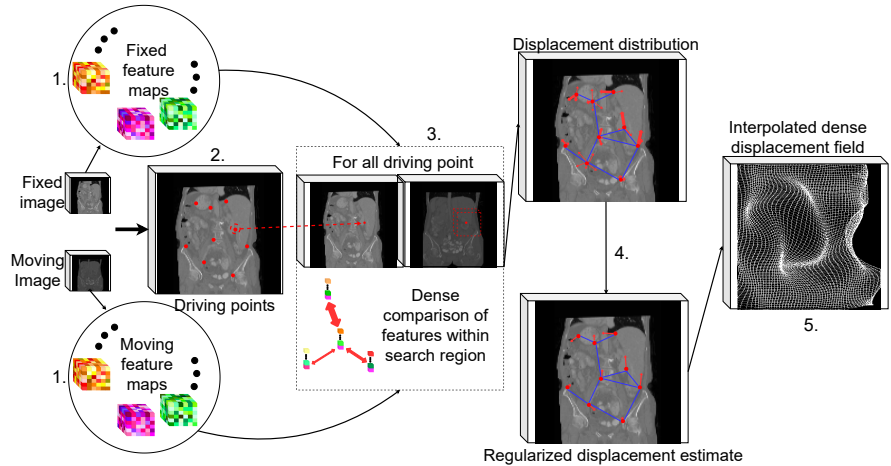}
\caption{Generic probabilistic displacement registration pipeline. \textit{1. Dense Feature Extraction}: Dense feature maps are extracted from both the moving and the fixed image. \textit{2. Driving Points Identification}: A subset of points from the fixed image are identified as driving points. \textit{3. Displacement Distribution Estimation}: A displacement distribution is obtained for each driving points by comparing its associated feature vector with and feature vectors from the moving feature maps within a predefined search region. \textit{4. Regularization}: a graph connecting the driving points is used to estimate a regularized displacement field. \textit{5. Interpolation}: the displacement field is interpolated to the original voxel grid.}
\label{FM pipeline generic}
\end{figure}

%
%
%
Medical image registration (MIR) is ubiquitous in medical engineering analytical pipelines. While there are many widely used MIR tools~\cite{Niftyreg,ANTs},
inter-patient abdominal registration
remains particularly challenging. The main challenge of this task is the high geometrical and appearance variabilities of the abdominal region of the human body. These result from the combination of intra-patient variability mainly due to the presence of soft organs highly impacted by the patient's position and current states (breathing, digestion, etc.) and inter-patient abdominal variance. 
Comparison studies~\cite{Xu2016,L2R} suggest two research directions for this task: first, Xu et al.~\cite{Xu2016} showed that probabilistic displacement methods tend to be less prone to local minima as opposed to other optimization-based MIR algorithms. Probabilistic displacement methods consider a probabilistic distribution over deformations derived from local points matching distribution as illustrated in Fig.~\ref{FM pipeline generic}. Secondly, the Learn2Reg (L2R) challenge~\cite{L2R} has demonstrated the superiority of learning-based approaches on this task. Indeed, learning-based regressive models such as LapIRN~\cite{LapIRN} have shown great success for inter-patient abdominal registration (L2R task 3). The learning-based counter-parts of probabilistic displacement approaches~\cite{Heinrich2019ClosingTG,Hansen2021.33872144} achieved great performances on US-MR brain registration, lung registration, and intra-patient abdominal image fusion but did not reach the same level of performance on inter-patient abdominal registration. Yet, these approaches allow for large deformations by design as they evaluate displacement probabilities within a (large) specified search region. Also, as probabilistic approaches, they should be more robust to the noise and ambiguity inherent to inter-patient abdominal registration.

Probabilistic displacement approaches typically have a high memory and computational cost due to the evaluation of displacement probabilities within a search region. Hence, these methods commonly operate on a subset of driving points (step 2. in Fig.~\ref{FM pipeline generic}) which are a lower resolution grid~~\cite{Glocker2008,HeinrichSimpson,Parisot2014,Heinrich2019ClosingTG}, uniformly sampled random points~\cite{Hansen2021} or key-points~\cite{10.1007/978-3-319-24571-3_41,Ruhaak2017,Hansen2021.33872144,Rister2017VolumetricIR,Shen2002,Xue2004}. These approaches disentangle the selection of the driving points from the rest of the pipeline. Yet, this step impacts the matching process (driving points should be matched unambiguously) as well as the deformation complexity (the driving points should be placed to capture the relevant variations of the deformation field). 
In this work, we proposed to learn a driving points predictor in an end-to-end fashion. As such, the driving points prediction is integrated and optimized alongside other learnable components and is specifically tailored to a particular probabilistic displacements registration pipeline. For natural image matching and perception, landmarks are usually obtained as local maxima of a response map~\cite{Thewlis_2017_ICCV,Zhang2018-cvpr-lmdis-rep}. This response map can be handcrafted or optimized as a neural network. We proposed instead to obtain the driving points by predicting a deformation field to be applied to a low-resolution regular grid. The vertices of the deformed regular grid are then considered as driving points. This format ensures a minimum coverage of the image and a fixed amount of driving points. It is also a memory-efficient prediction format that still allows predicting locations with sub-voxel precision. We conducted experiments on two datasets, a CT dataset from the L2R 2020 challenge~\cite{Xu2016} and the CHAOS MR dataset~\cite{CHAOSdata2019}. We demonstrate improvement in most probabilistic displacement models when using a driving points predictor against two other standard driving points selection methods.

\section{Method}
In this section, we first describe the five steps of probabilistic displacement registration pipelines (see Fig.~\ref{FM pipeline generic}) and their associated methods from the literature before introducing in detail the proposed driving points prediction method which can be used for the second step to improve the overall pipeline performances. 

\subsection{Pipeline overview}\label{background}

\paragraph{1. Dense Feature Extraction}\label{Feature Extraction}
%
The objective of the dense feature extractor is to associate a vector characterizing the corresponding anatomical point to each voxel. 
Various handcrafted feature extraction methods have been suggested. The most straightforward option is to rely on image intensities~\cite{Foroughi2005,Glocker2008,garcia:inria-00616148,Heinrich2013.02780062,Castillo2014} and/or intensity gradients~\cite{Foroughi2005,HeinrichSimpson,Ruhaak2017}. Wavelet decomposition~\cite{Xue2004} and 3D Gabor attributes~\cite{Ou2011} have been used to enrich these feature vectors. Other methods proposed to describe voxels using local self-similarity measurements such as the self-similarity context~\cite{Heinrich2013} or MIND descriptors~\cite{10.1007/978-3-319-24571-3_41,Hansen2021.33872144,Hansen2021}. Finally, recent models adopted a data-driven approach by learning deep feature extractors~\cite{Heinrich2019ClosingTG,HeinrichHanser,Ha2019}.

\paragraph{2. Driving Points Identification}\label{Driving Points Identification}

Instead of matching all points, probabilistic displacement models typically focus on a subset of points called driving points. This helps to reduce the computational cost. If driving points are well selected, it also mitigates the impact of points located in ambiguous regions and, thus, difficult to match accurately.
The most straightforward approach is to rely on a low resolution regular grid~\cite{Glocker2008,Glocker2011,HeinrichSimpson,Parisot2014,Heinrich2019ClosingTG} or to use random points drawn from a uniform distribution~\cite{Hansen2021}. More sophisticated approaches rely on key-points  extractors such as the Foerstner key-point extractor~\cite{10.1007/978-3-319-24571-3_41,Ruhaak2017,Hansen2021.33872144}, 3D-SIFT points~\cite{Rister2017VolumetricIR} or handcrafted key-points extractors~\cite{Shen2002,Yap2009,Xue2004}. Key-points extractor based on the image input only (no additional segmentation provided) focus on salient regions in the fixed image by considering local maxima of a saliency function. Yet, for abdominal image MIR, this does not necessarily mean that the point can be unambiguously matched or is of interest for the deformation. Another drawback of such methods is the lack of guarantee to cover sufficiently the regions of interest in the image. In contrast, the proposed driving points predictor presented in the following section covers by design a large region of the abdomen and optimizes a typical registration objective term.

\paragraph{3. Displacement Distribution Estimation}\label{Displacement Distribution Estimation}

The next step estimates a displacement distribution for each driving point. Specifically, a matching probability is derived from the Euclidean distance between the feature vectors describing each driving point and the feature vectors of matching candidates. In general, matching candidates are assumed to be in a pre-defined window centered around the driving point~\cite{Heinrich2019ClosingTG,Hansen2021.33872144}. Alternatively, random displacements sampled from uniform distributions can be used~\cite{Heinrich2013.02780062,Hansen2021}.

\paragraph{4. Regularization}\label{Regularization and Unification}

The regularization step promotes smooth deformations, which are assumed to be more realistic. In order to promote spatially coherent deformations, several options have been proposed such as fitting a transformation model having low degree of freedom to the matching distribution~\cite{Castillo2014}. Regarding probabilistic regularization models, deformations are typically assumed to be drawn from a Markov Random Field (MRF) distribution. The individual displacement distributions estimated at the previous step define the unary term of the MRF distribution, while the pairwise term promote local displacement coherence. The maximum a posteriori (MAP) estimate of this distribution can be computed using the Fast-PD algorithm~\cite{Glocker2008,Glocker2011,Ou2011,Parisot2014}. As this algorithm introduces a high memory and computational cost, the resolution of the MRF distribution can be approximated by replacing the neighboring graph with a minimum spanning tree to compute the MAP estimator in a closed form~\cite{HeinrichSimpson,Heinrich2013.02780062,Heinrich2013,10.1007/978-3-319-24571-3_41,Ruhaak2017}. Alternatively, Mean Field (MF) approximation can be employed to approximate the mean estimator or the MAP estimator~\cite{Heinrich2019ClosingTG,HeinrichHanser,Hansen2021}. Finally, regularization can be performed using a neural network~\cite{Ha2019,Hansen2021.33872144}.

\paragraph{5. Interpolation}\label{interpolation}

The last step consists of interpolating the deformation back to the original resolution. Standard approaches rely on non-overlapping~\cite{Shen2002,Xue2004,Foroughi2005} and overlapping~\cite{garcia:inria-00616148} Gaussian kernel interpolation or linear interpolation~\cite{Heinrich2019ClosingTG}.

\subsection{Driving Points Prediction}

Driving point sets should correspond to points which can be reliably matched within their associated search region in the moving image. For this reason, the driving points selection should be adapted to the feature extraction step. Ideally, the driving point set should be specific to a fixed/moving image pair rather than only depending on the fixed image as the ambiguity of the matching process depends on both inputs. 
Finally, the driving points selection impacts the subsequent steps of the pipeline so it should be adapted to the pipeline as a whole, just like the pipeline should be adapted to it. For this reason, we propose to learn the driving points selection process in an end-to-end fashion to select points that are adapted to the rest of the pipeline to optimize the registration objective. 
\begin{figure}[t]
\includegraphics[width=\linewidth]{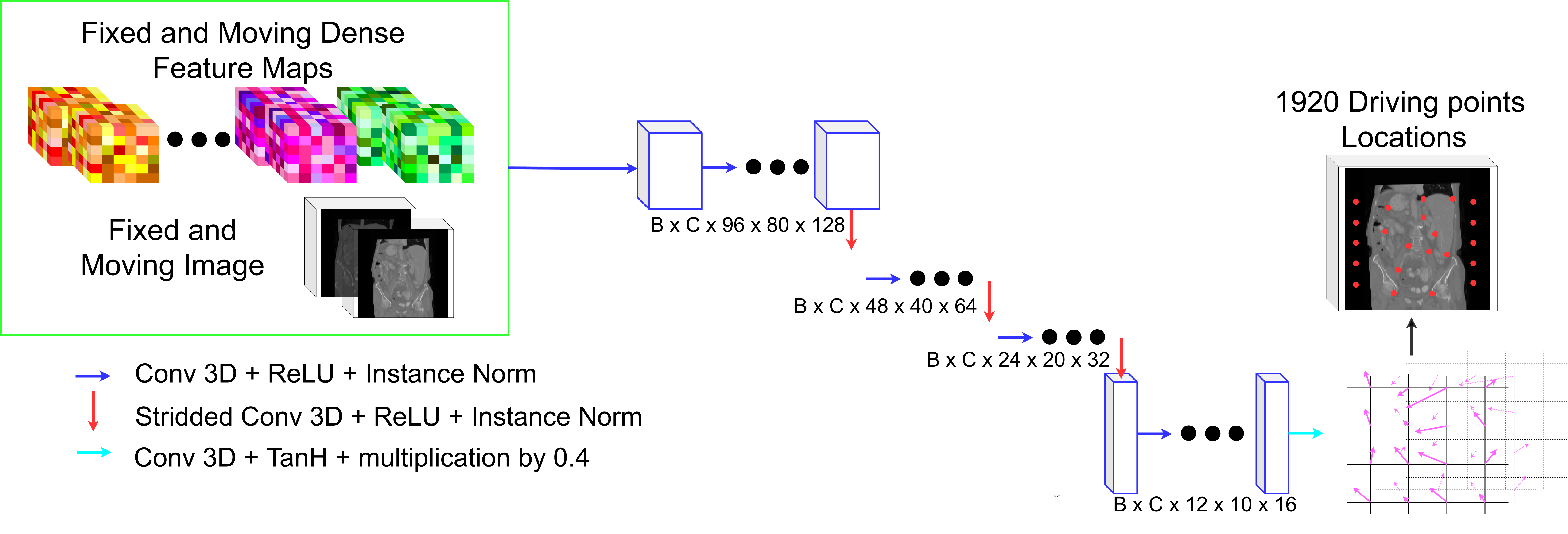}
\caption{Driving points predictor architecture}
\label{Driving points prediction}
\end{figure}

The proposed architecture for the driving points predictor is illustrated in Fig.~\ref{Driving points prediction}. The driving points predictor takes as input the concatenation of both the fixed and the moving image as well as their associated dense feature maps. Hence, the driving points predictor can exploit the specific feature maps landscape to predict relevant points to match.
The input tensor goes through an encoder network with several downsampling layers to obtain a lower resolution embedding of the input tensor. From there, directly predicting the driving points' locations (by flattening the low-resolution embedding and passing it through a set of fully connected layers for instance) is particularly challenging as it is an unstructured prediction process very sensitive to the random initialization. Instead, we propose the use of a structured output by deforming a low-resolution regular grid. Indeed, as illustrated in Fig.~\ref{Driving points prediction}, the final layer of the driving points predictor is also convolutional and predicts a displacement field to be applied to the low-resolution regular grid. The vertices of the deformed low-resolution grid are considered as driving points. To ensure a minimum coverage of the image, we cap the maximum displacement range to $20\%$ of the image dimension in all directions. The number of predicted driving points typically corresponds to the number $G$ of vertices in the low-resolution regular grid. In theory, multiple vertices could be mapped to the same points, leading to fewer driving points but we never observed that phenomenon in practice. In case more driving points are required to improve performances, one could either use a higher resolution grid (i.e. use fewer downsampling layers) or make the last layer predict $D$ different displacement fields to obtain $D \times G$ driving points.

Once the driving points' locations have been predicted, the fixed dense feature map is sampled at the predicted locations using linear interpolation to obtain the descriptors of the driving points. This step is differentiable and allows to learn the driving points predictor parameters through back-propagation. The parameters of the driving points predictor are learned simultaneously with other learnable components of the pipeline (e.g. the feature extractor or the regularization network) by minimizing a standard registration loss, which in our experiments is a combination of local normalized cross-correlation and a Hessian penalty term (mathematical formulation in supplementary material).



\section{Experiments}

\subsection{Datasets}
We conducted experiments on two different datasets. First, we used the combined training and validation sets of the L2R 2020 challenge task 3~\cite{L2R}. This dataset contains 30 Abdominal CT volumes. These volumes also come with the segmentation of 13 structures: the spleen, the left/right kidney, the gallbladder, the esophagus, the liver, the stomach, the aorta, the inferior vena cava, the portal and splenic veins, the pancreas, the left/right adrenal gland. Then, we considered the Magnetic Resonance (MR) volumes of the CHAOS dataset~\cite{CHAOSdata2019}. This dataset contains 40 MR volumes from different patients with 4 structures annotated: the spleen, the left/right kidney, and the liver. For each dataset, we saved 10 volumes for testing (i.e. 90 pairs of patients to be registered) and used the rest of the volumes for training/validation. We refer from now on to the L2R challenge dataset as "CT dataset" and to the Chaos dataset as "MR dataset".

\subsection{Experimental setting}

As detailed in Section~\ref{background}, various solutions have been proposed for each step. In this work, we introduced a new data-driven and feature-aware driving points selection process. To assess the effectiveness of the proposed method, we learned driving points predictors as part of different pipelines. Specifically, we considered three different dense feature extractors: the image intensity as a baseline, MIND descriptors~\cite{10.1007/978-3-319-24571-3_41,Hansen2021.33872144,Hansen2021} as a state-of-the-art option for handcrafted features for MIR or feature vectors extracted using a standard UNet architecture~\cite{UNet}. We then compared our driving points predictor with two traditional driving points selection approaches based on a low-resolution grid and Foerstner key-points. Regarding the displacement distribution estimation, we used a predefined search region associated to the set of admissible displacements $\{-25, -20, ... 20, 25\}^3$ (expressed in voxels). Two forms of regularization were tested, either based on mean-field approximation of an MRF formulation as in PDDNet~\cite{Heinrich2019ClosingTG} or a neural network graph-based approach as in GraphRegNet (GRN)~\cite{Hansen2021.33872144}. Linear interpolation was used for the interpolation step.

To assess the quality of the registration, we report Dice scores for each annotated structure. We also report regularity metrics in the supplementary material. All learnable components were trained in an end-to-end fashion by minimizing a standard MIR loss combining the local normalized cross-correlation loss as similarity metric and a regularization term penalizing the norm of the Hessian of the displacement field. Not using label information during training allows a fair comparison with methods that have very limited learned parameters.  All hyper-parameters were optimized using grid-search and are available in our code\footnote{\href{https://github.com/SamuelJoutard/DrivingPointsPredictionMIR}{https://github.com/SamuelJoutard/DrivingPointsPredictionMIR}}. 

\begin{figure}[t]
\centering
     \begin{subfigure}[b]{0.48\linewidth}
         \centering
         \includegraphics[width=\linewidth]{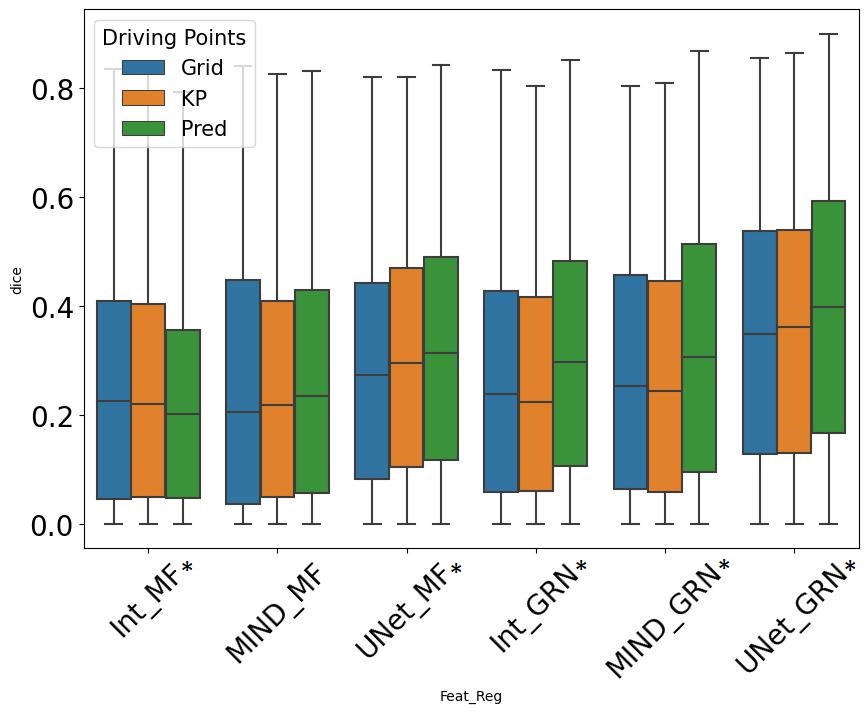}
         \caption{CT dataset}
         \label{Mean Dice L2R}
     \end{subfigure}
    \centering
     \begin{subfigure}[b]{0.48\linewidth}
         \centering
         \includegraphics[width=\linewidth]{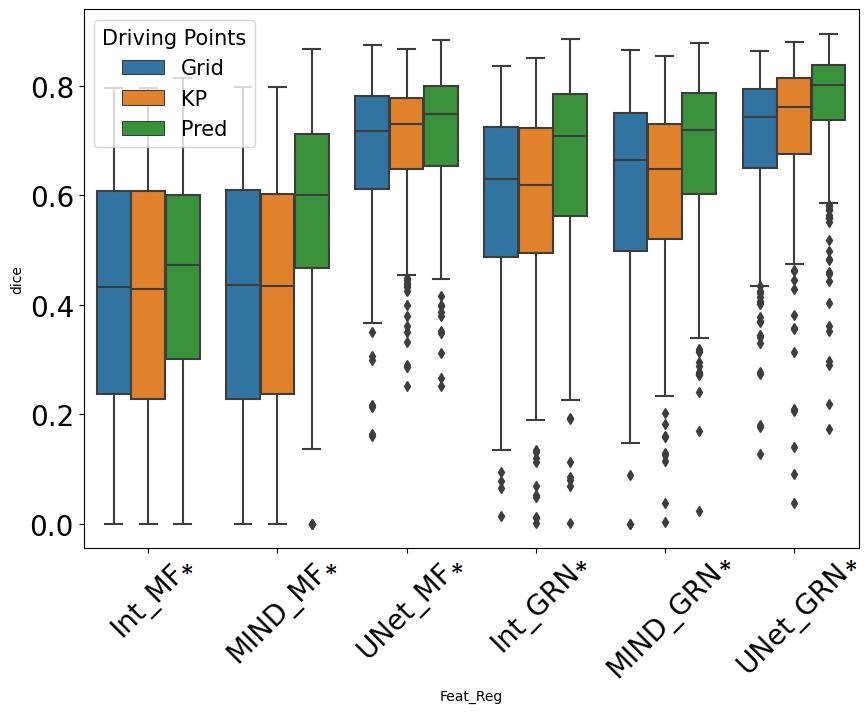}
         \caption{MR dataset}
         \label{Mean Dice CHAOS}
     \end{subfigure}
     \hfill
\caption{Mean Dice scores for the considered probabilistic displacment registration pipelines. Grid, KP and Pred respectively correspond to using a low resolution grid, Foerstner Key-points and the proposed driving points predictor. The x-axis indicates the feature/regularization configuration. (*) indicates configurations where the ranking is statistically significant (Wilcoxon rank test, $p<0.01$).}
\label{Mean dice}
\end{figure}

\subsection{Results}

Averaged Dice scores are reported in Fig.~\ref{Mean dice} and more detailed results are available in Tables~\ref{Exp1} and~\ref{Exp2} of supplementary material. Our method significantly improved the registration compared to using Foerstner key-points or a low resolution grid for 10 of the 12 Features/Regularization/Dataset configurations tested. This highlights the effectiveness of our approach. The improvement observed for the MIND/Mean-Field/CT dataset is not statistically significant by the Wilcoxon test and we observed a performance downgrade for the intensity/mean-field/CT dataset due to some overfitting we did not manage to prevent.
We also note that our approach leads to large improvements on the MR dataset when using the intensity or MIND features which shows that optimizing the driving points for a specific dense feature extraction method is particularly effective when the feature extraction step is sub-optimal. The small structures segmented in the CT dataset such as the gallbladder are particularly challenging to align which explains the lower mean Dice scores observed on this dataset.

\begin{figure}[t]
\centering
     \begin{subfigure}[b]{0.19\linewidth}
         \centering
         \includegraphics[width=\linewidth]{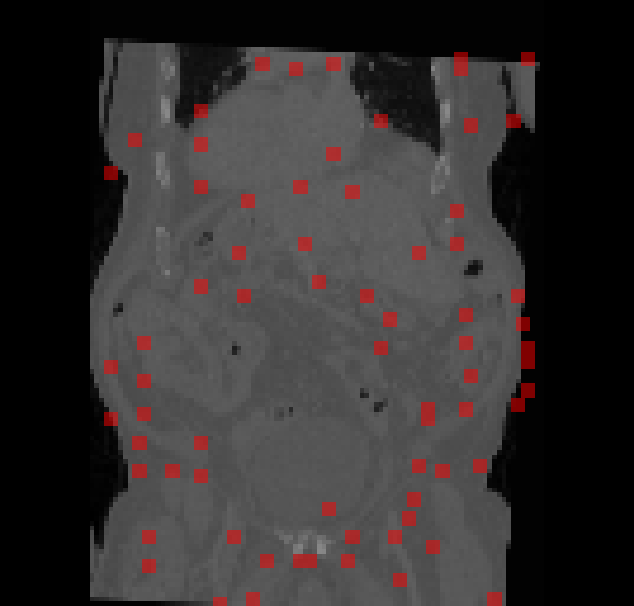}
         \caption{}
         \label{a}
     \end{subfigure}
    \centering
     \begin{subfigure}[b]{0.19\linewidth}
         \centering
         \includegraphics[width=\linewidth]{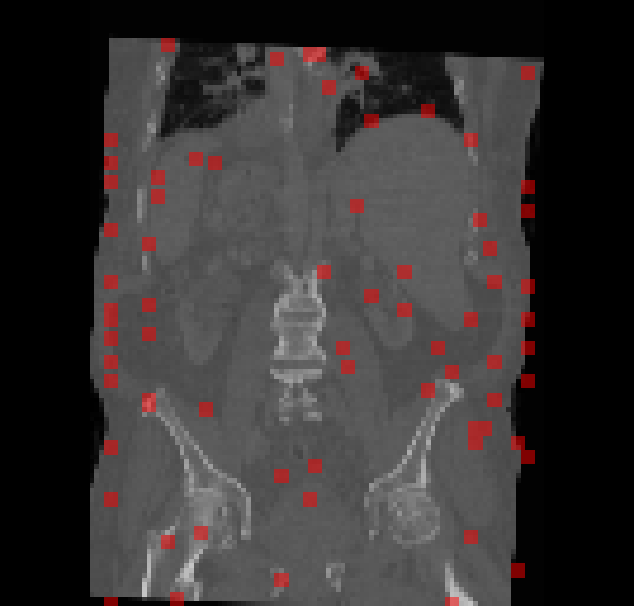}
         \caption{}
         \label{b}
     \end{subfigure}
    \centering
     \begin{subfigure}[b]{0.19\linewidth}
         \centering
         \includegraphics[width=\linewidth]{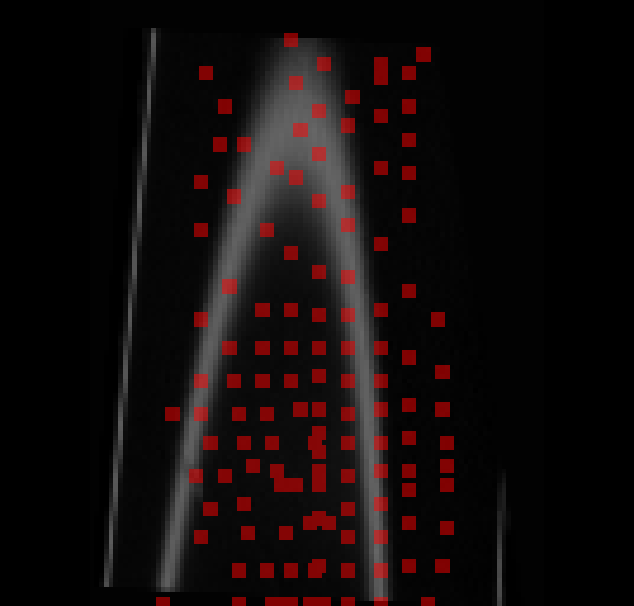}
         \caption{}
         \label{c}
     \end{subfigure}
    \centering
     \begin{subfigure}[b]{0.19\linewidth}
         \centering
         \includegraphics[width=\linewidth]{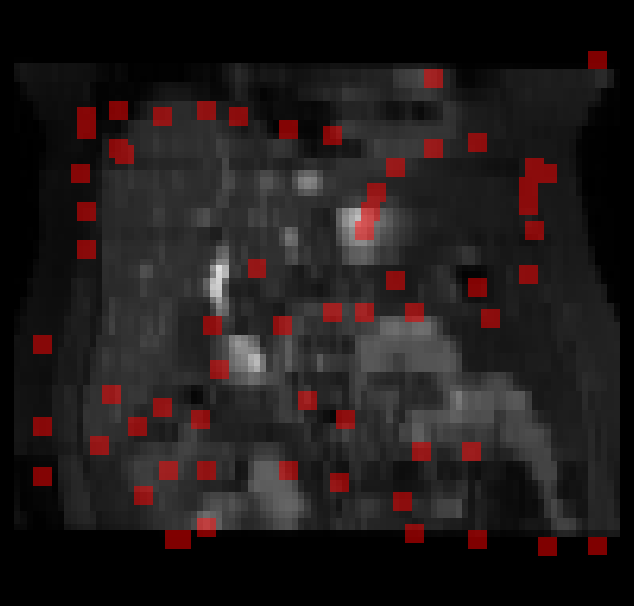}
         \caption{}
         \label{d}
     \end{subfigure}
    \centering
     \begin{subfigure}[b]{0.19\linewidth}
         \centering
         \includegraphics[width=\linewidth]{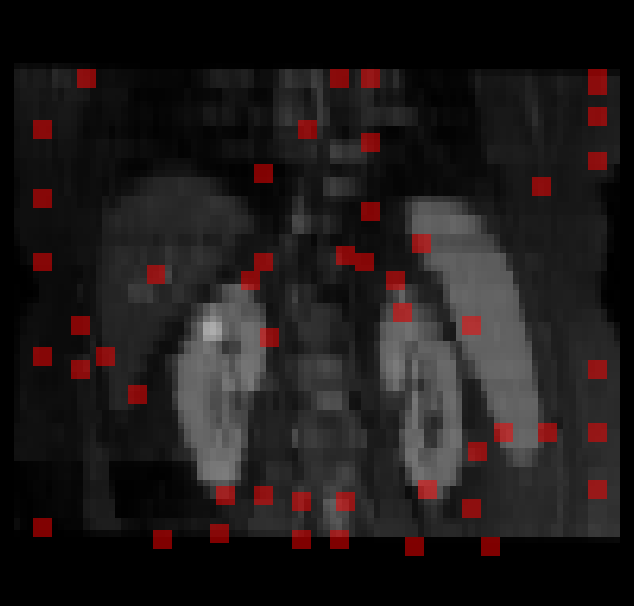}
         \caption{}
         \label{e}
     \end{subfigure}
     \hfill
 \caption{Predicted driving points represented in red overlapping the fixed image. (a,b,c) show coronal slices of the same patient from the CT dataset. (c) shows driving points predicted far from the regions of interest. (d,e) show coronal slices of the same patient from the MR dataset.}
 \label{Driving points examples}
\end{figure}

We illustrate some of the predicted driving points in Fig.~\ref{Driving points examples}. First, we observe by comparing Fig.~\ref{c} to the other slices shown that the regular grid structure is only preserved far from the regions of interest. The driving points predictor selected specific locations within the fixed patient's abdomen. We can also observe that regions that are hard to match like the intestinal region (see Fig.~\ref{a}) or regions with low contrast like the interior of the liver (Fig.~\ref{b}~\ref{d}) are not densely populated with driving points. This suggests that the registration pipeline tends to focus on informative points within the patient when using our method.

Finally, to evaluate the respective contributions of the fixed and moving images in the driving points selection, we compared predicted driving point sets. Specifically, we computed the Wassertein-2 distance (W2) between predicted driving point sets for different inputs. We compared the mean W2 distance between predicted driving point sets used to register two pairs of images sharing the same fixed image against the mean W2 distance between all predicted driving point sets. The former is on average 7 times smaller than the latter. This suggests that the predicted driving point sets are specific to the fixed image. More detailed comparisons are available in Fig.\ref{Mean W2} of supplementary materials. We also compared predicted driving points for the same input pair of images to be registered but using different driving points predictors trained for different pipelines. We obtained mean W2 distance that ranges from 13 to 41 for the CT dataset and from 11 to 29 for the MR dataset. This confirms that the predicted driving point sets are pipeline specific. 

\section{Discussion}

In this work, we proposed an encoder architecture to predict driving point set for probabilistic displacements MIR models applied to abdominal images. The proposed architecture is stable to train, ensures a certain coverage of the fixed image, and has a memory-saving prediction format. We showed that predicting driving points in this way yielded performance improvements for almost all configurations tested which involved two different datasets. Finally, we showed that the proposed architecture was able to predict driving point sets tailored to the fixed image data and to the specific steps of the registration pipeline. 


The presented methods did not integrate all the potential refinements that have been shown to improve performances such as multiple forward passes with intermediate wrapping or instance optimization. Similarly, in this work, the segmentation masks were used for evaluation purposes only. Given that previous studies have shown that exploiting these segmentation masks during training leads to performance improvement, we plan to integrate a segmentation matching loss alongside other model improvements in future work to further improve our results.

\bibliographystyle{splncs04}
\bibliography{ref.bib}

\begin{thebibliography}{10}
\providecommand{\url}[1]{\texttt{#1}}
\providecommand{\urlprefix}{URL }
\providecommand{\doi}[1]{https://doi.org/#1}

\bibitem{Castillo2014}
Castillo, E., Castillo, R., Fuentes, D., Guerrero, T.: Computing global
  minimizers to a constrained b-spline image registration problem from optimal
  l 1 perturbations to block match data  (2014)

\bibitem{Foroughi2005}
Foroughi, P., Abolmaesumi, P.: Elastic registration of 3d ultrasound images.
  Medical image computing and computer-assisted intervention : MICCAI ...
  International Conference on Medical Image Computing and Computer-Assisted
  Intervention  (2005)

\bibitem{garcia:inria-00616148}
Garcia, V., Commowick, O., Malandain, G.: {A Robust and Efficient Block
  Matching Framework for Non Linear Registration of Thoracic CT Images}. In:
  {Grand Challenges in Medical Image Analysis (MICCAI workshop)} (2010)

\bibitem{Glocker2008}
Glocker, B., Komodakis, N., Tziritas, G., Navab, N., Paragios, N.: Dense image
  registration through mrfs and efficient linear programming. Medical Image
  Analysis  (2008)

\bibitem{Glocker2011}
Glocker, B., Sotiras, A., Komodakis, N., Paragios, N.: Deformable medical image
  registration: Setting the state of the art with discrete methods. Annual
  Review of Biomedical Engineering  (2011)

\bibitem{Ha2019}
Ha, I.Y., Heinrich, M.P.: Comparing deep learning strategies and attention
  mechanisms of discrete registration for multimodal image-guided interventions
  (2019)

\bibitem{Hansen2021.33872144}
Hansen, L., Heinrich, M.P.: Graphregnet: Deep graph regularisation networks on
  sparse keypoints for dense registration of 3d lung cts  (2021)

\bibitem{Hansen2021}
Hansen, L., Heinrich, M.P.: Revisiting iterative highly efficient optimisation
  schemes in medical image registration  (2021)

\bibitem{Heinrich2019ClosingTG}
Heinrich, M.P.: Closing the gap between deep and conventional image
  registration using probabilistic dense displacement networks. In: MICCAI
  (2019)

\bibitem{10.1007/978-3-319-24571-3_41}
Heinrich, M.P., Handels, H., Simpson, I.J.A.: Estimating large lung motion in
  copd patients by symmetric regularised correspondence fields. In: Medical
  Image Computing and Computer-Assisted Intervention -- MICCAI 2015 (2015)

\bibitem{HeinrichHanser}
Heinrich, M.P., Hansen, L.: Highly accurate and memory efficient unsupervised
  learning-based discrete ct registration using 2.5d displacement search

\bibitem{Heinrich2013.02780062}
Heinrich, M.P., Jenkinson, M., Brady, M., Schnabel, J.A.: Mrf-based deformable
  registration and ventilation estimation of lung ct. IEEE Transactions on
  Medical Imaging  (2013)

\bibitem{HeinrichSimpson}
Heinrich, M.P., Simpson, I.J.A., Jenkinson, M., Brady, M., Schnabel, J.A.:
  Uncertainty estimates for improved accuracy of registration-based
  segmentation propagation using discrete optimisation

\bibitem{Heinrich2013}
Heinrich, M.P., Jenkinson, M., 1, B.W.P.Z., Brady, M., Schnabel, J.A.: Towards
  realtime multimodal fusion for image-guided interventions using
  self-similarities  (2013)

\bibitem{L2R}
Hering, A., Hansen, L., Mok, T.C.W., Chung, A.C.S., Siebert, H., Häger, S.,
  Lange, A., Kuckertz, S., Heldmann, S., Shao, W., Vesal, S., Rusu, M., Sonn,
  G., Estienne, T., Vakalopoulou, M., Han, L., Huang, Y., Brudfors, M.,
  Balbastre, Y., Joutard, S., Modat, M., Lifshitz, G., Raviv, D., Lv, J., Li,
  Q., Jaouen, V., Visvikis, D., Fourcade, C., Rubeaux, M., Pan, W., Xu, Z.,
  Jian, B., De~Benetti, F., Wodzinski, M., Gunnarsson, N., Sjölund, J., Qiu,
  H., Li, Z., Großbröhmer, C., Hoopes, A., Reinertsen, I., Xiao, Y., Landman,
  B., Huo, Y., Murphy, K., Lessmann, N., van Ginneken, B., Dalca, A.V.,
  Heinrich, M.P.: Learn2reg: comprehensive multi-task medical image
  registration challenge, dataset and evaluation in the era of deep learning
  (2021)

\bibitem{CHAOSdata2019}
Kavur, A.E., Selver, M.A., Dicle, O., Barış, M., Gezer, N.S.: {CHAOS -
  Combined (CT-MR) Healthy Abdominal Organ Segmentation Challenge Data} (2019)

\bibitem{Niftyreg}
Modat, M., Ridgway, G., Taylor, Z., Lehmann, M., Barnes, J., Hawkes, D., Fox,
  N., Ourselin, S.: Fast free-form deformation using graphics processing units.
  Computer methods and programs in biomedicine  (2009)

\bibitem{LapIRN}
Mok, T.C.W., Chung, A.C.S.: Large deformation diffeomorphic image registration
  with laplacian pyramid networks (2020)

\bibitem{Ou2011}
Ou, Y., Sotiras, A., Paragios, N., Davatzikos, C.: Dramms: Deformable
  registration via attribute matching and mutual-saliency weighting. Medical
  Image Analysis  (2011)

\bibitem{Parisot2014}
Parisot, S., Wells, W., Chemouny, S., Duffau, H., Paragios, N.: Concurrent
  tumor segmentation and registration with uncertainty-based sparse non-uniform
  graphs. Medical Image Analysis  (2014)

\bibitem{Rister2017VolumetricIR}
Rister, B., Horowitz, M., Rubin, D.: Volumetric image registration from
  invariant keypoints. IEEE Transactions on Image Processing  (2017)

\bibitem{UNet}
Ronneberger, O., Fischer, P., Brox, T.: U-net: Convolutional networks for
  biomedical image segmentation. In: Navab, N., Hornegger, J., Wells, W.M.,
  Frangi, A.F. (eds.) Medical Image Computing and Computer-Assisted
  Intervention -- MICCAI 2015 (2015)

\bibitem{Ruhaak2017}
Ruhaak, J., Polzin, T., Heldmann, S., Simpson, I.J., Handels, H., Modersitzki,
  J., Heinrich, M.P.: Estimation of large motion in lung ct by integrating
  regularized keypoint correspondences into dense deformable registration. IEEE
  Transactions on Medical Imaging  (2017)

\bibitem{Shen2002}
Shen, D., Davatzikos, C.: Hammer: Hierarchical attribute matching mechanism for
  elastic registration  (2002)

\bibitem{Thewlis_2017_ICCV}
Thewlis, J., Bilen, H., Vedaldi, A.: Unsupervised learning of object landmarks
  by factorized spatial embeddings. In: Proceedings of the IEEE International
  Conference on Computer Vision (ICCV) (Oct 2017)

\bibitem{ANTs}
Tustison, N.J., Cook, P.A., Klein, A., Song, G., Das, S.R., Duda, J.T., Kandel,
  B.M., {van Strien}, N., Stone, J.R., Gee, J.C., Avants, B.B.: Large-scale
  evaluation of ants and freesurfer cortical thickness measurements. NeuroImage
   (2014)

\bibitem{Xu2016}
Xu, Z., Lee, C.P., Heinrich, M.P., Modat, M., Rueckert, D., Ourselin, S.,
  Abramson, R.G., Landman, B.A.: Evaluation of six registration methods for the
  human abdomen on clinically acquired ct. IEEE Transactions on Biomedical
  Engineering  (2016)

\bibitem{Xue2004}
Xue, Z., Shen, D., Davatzikos, C.: Determining correspondence in 3-d mr brain
  images using attribute vectors as morphological signatures of voxels. IEEE
  Transactions on Medical Imaging  (2004)

\bibitem{Yap2009}
Yap, P.T., Wu, G., Zhu, H., Lin, W., Shen, D.: Timer: Tensor image morphing for
  elastic registration. NeuroImage  (2009)

\bibitem{Zhang2018-cvpr-lmdis-rep}
Zhang, Y., Guo, Y., Jin, Y., Luo, Y., He, Z., Lee, H.: Unsupervised discovery
  of object landmarks as structural representations. In: Conference on Computer
  Vision and Pattern Recognition ({CVPR}) (2018)

\end{thebibliography}

\pagebreak
\begin{center}
\textbf{\large Supplemental Materials: Driving Points Prediction For Abdominal Probabilistic Registration}
\end{center}
\setcounter{section}{0}
\setcounter{equation}{0}
\setcounter{figure}{0}
\setcounter{table}{0}
\setcounter{page}{1}
\makeatletter
\renewcommand{\theequation}{S\arabic{equation}}
\renewcommand{\thefigure}{S\arabic{figure}}
\renewcommand{\thetable}{S\arabic{table}}





\section{Mathematical formulations}

\subsection{Registration}
We specify here the rigorous mathematical formulation of the registration objective as mentioned in Sections 2.2 and 3.2.

When registering two images non-rigidly, we typically combine a similarity and a regularization term. The first one encourages content alignment while the second encourages smooth deformations. Hence, given a fixed image $\mathcal{F}:\Omega\longrightarrow \mathbb{R}$ and a moving image $\mathcal{M}:\Omega\longrightarrow \mathbb{R}$ ($\Omega$ is the spatial domain of the image), a registration algorithm is looking for a deformation field $\phi: \Omega \longrightarrow \Omega$ so that:

\begin{equation}\label{Reg obj}
    \phi = \argmin_\phi \mathcal{S}(\mathcal{F}, \mathcal{M} \circ \phi) + \lambda R(\phi)
\end{equation}

When considering a learning approach, we optimize the parameter set $\beta$ of a function $f: \mathcal{M}, \mathcal{F}, \beta \longrightarrow \phi$ that outputs a deformation field given the two images to be registered (and the function's parameter set). In that case, the parameter set minimises the objective~\ref{Reg obj} in expectancy for pairs of fixed/moving images sampled from a training distribution $\mathcal{T}$. Hence, $\beta$ is obtained so that:

\begin{equation}
    \beta = \argmin_\beta \mathbb{E}_{(\mathcal{M},\mathcal{F})\sim\mathcal{T}}\left[\mathcal{S}\left(\mathcal{F}, \mathcal{M} \circ f(\mathcal{M}, \mathcal{F}, \beta)\right) + \lambda R\left(f(\mathcal{M}, \mathcal{F}, \beta)\right)\right]
\end{equation}

In our case, $\mathcal{S}$ is the local normalised cross correlation and $R$ is the bending energy respectively given by Equations~\ref{lncc} and~\ref{bending energy}. We only provide these formula for discrete images as these are what we manipulate in practice. 

\paragraph{Local Normalised Cross Correlation}

\begin{equation}\label{lncc}
    \mathcal{S}(\mathcal{I}_1, \mathcal{I}_2) = \sum\limits_{p\in \Omega} \frac{\sum\limits_{v\in\mathcal{V}} \left(\mathcal{I}_1(p+v) - \Tilde{\mathcal{I}}_1(p)\right) \left(\mathcal{I}_2(p+v) - \Tilde{\mathcal{I}}_2(p)\right)}{\sqrt{\left(\sum\limits_{v\in\mathcal{V}} \left(\mathcal{I}_1(p+v) - \Tilde{\mathcal{I}}_1(p)\right)^2\right) \left(\sum\limits_{v\in\mathcal{V}} \left(\mathcal{I}_2(p+v) - \Tilde{\mathcal{I}}_2(p)\right)^2\right)}}
\end{equation}
Where $\mathcal{I}_1$ and $\mathcal{I}_2$ are two images and $\mathcal{V}$ is a set of offsets defining a neighborhood. $\Tilde{\mathcal{I}}_*(p)$ is the local intensity mean of the image $\mathcal{I}_*$ i.e. $\Tilde{\mathcal{I}}_*(p)=\sum\limits_{v \in \mathcal{V}}\mathcal{I}_*(p+v)$.

\paragraph{Bending Energy}

\begin{equation}\label{bending energy}
    R(\phi) = \sum\limits_{p \in \Omega} \sum\limits_{i,j\in\{1, 2, 3\}}\Delta_i\Delta_j\phi (p)^T\Delta_i\Delta_j\phi (p)
\end{equation}

Where $\Delta_i$ is the discrete differential operator along the direction $i$ defined as:

\begin{equation}\label{bending energy}
    \Delta_i \phi (p) = \phi(p + e_i) - \phi(p)
\end{equation}

Where $e_i$ is the unit coordinate vector along the $i$ axis.

The local normalised cross-correlation encourages aligned patches to have correlated intensities while the bending energy penalises second order variations of the deformation field. 

\subsection{Markov Random Field for Probabilistic Displacement methods}

In this section we explicit the probabilistic formulation commonly used in probabilistic displacement models. This formulation will cover the processing step from the step 1. (dense feature extraction) to the step 4. (Regularization) of the pipeline presented in the main text.

As in the previous section, let $\mathcal{F}:\Omega\longrightarrow \mathbb{R}$ be the fixed image and $\mathcal{M}:\Omega\longrightarrow \mathbb{R}$ be to moving one ($\Omega$ is the spatial domain of the images). During step 1., dense feature maps $F_{\mathcal{F}}:\Omega\longrightarrow \mathbb{R}^d$ and $F_{\mathcal{M}}:\Omega\longrightarrow \mathbb{R}^d$ are respectively extracted from the fixed and the moving image. $d$ is the dimension of the feature vectors associated to each voxel. Then, during step 2., a set of driving points $\mathcal{O} \subset \Omega$ are identified to drive the registration process. For all driving points, we compute a matching potentials across a set of admissible displacements $\mathcal{D}$. For instance, deriving matching potentials from cosine similarity we obtain:

\begin{equation}
    \forall (p,\delta)\in \mathcal{O}\times\mathcal{D} \; \; \mu_{p,\delta} = \frac{F_{\mathcal{F}}(p)^T F_{\mathcal{M}}(p+\delta)}{\sqrt{\norm{F_{\mathcal{F}}(p)}_2^2\norm{F_{\mathcal{M}}(p+\delta)}_2^2}}
\end{equation}

A neighboring graph is then created to connect the set of driving points $\mathcal{O}$. Hence, for each driving point $p\in \mathcal{O}$, there is a set of neighbors $\mathcal{N}_p\subset\mathcal{O}$ associated to $p$ according to the neighboring graph.
A distribution with density $\mathcal{P}$ over displacement fields $\left.\psi\right|_{\mathcal{O}}$ restrained to the set of driving points with admissible displacements i.e. $\left.\psi\right|_{\mathcal{O}} = \left.(\phi - Id)\right|_{\mathcal{O}}: \mathcal{O}\longrightarrow \mathcal{D}$ is then obtained according to the following Markov Random Field formulation:

\begin{equation}
    log(\mathcal{P}(\psi)) \propto \sum\limits_{p\in \mathcal{O}} \mu_{p,\psi(p)} - \lambda \sum\limits_{p\in \mathcal{O}}\sum\limits_{p_n\in \mathcal{N}_p} exp\left(-\frac{\norm{p - p_n}_2^2}{2\sigma_p} \right )\norm{\psi(p) - \psi(p_n)}_2^2 
\end{equation}

Where $\lambda$ is a weighting factor and $\sigma_p$ is a spatial bandwidth within which displacements are encouraged to be correlated. This formulation is then either exploited as is or approximated via Mean-Field inference as in our experiments. 

\section{Additional Results and Figures}

\begin{landscape}

\begin{table}[H]
    \centering
    \begin{tabular}{|c|c|c||c|c|c|c|c|c|c|c|c|c|c|c|c|c|}
        \hline
         \multicolumn{3}{|c|}{Configuration choice} &  \multicolumn{14}{|c|}{Dice}\\
         \hline
         \rotatebox[origin=c]{70}{Features} & \rotatebox[origin=c]{70}{Driving points} & \rotatebox[origin=c]{70}{Regularization} &  \crule[green]{0.3cm}{0.3cm} & \crule[red]{0.3cm}{0.3cm} & \crule[blue]{0.3cm}{0.3cm} & \crule[yellow]{0.3cm}{0.3cm} & \crule[purple]{0.3cm}{0.3cm} & \crule[pink]{0.3cm}{0.3cm} &\crule[orange]{0.3cm}{0.3cm} & \crule[brown]{0.3cm}{0.3cm} & \crule[cyan]{0.3cm}{0.3cm} &\crule[magenta]{0.3cm}{0.3cm} & \crule[olive]{0.3cm}{0.3cm} & \crule[teal]{0.3cm}{0.3cm} & \crule[gray]{0.3cm}{0.3cm} & mean\\
         \hline
         \hline

\multirow{3}{*}{Intensity} & Grid & \multirow{3}{*}{Mean Field} & \textbf{41.4} & \textbf{33.7} & \textbf{34.6} & 2.0 & \textbf{22.0} & \textbf{62.5} & \textbf{24.0} & \textbf{32.7} & \textbf{36.0} & 4.9 & \textbf{15.2} & 7.6 & 8.4 & \textbf{25.0}\\
\cline{4-17}
 & Key-points &  & 40.8 & 33.3 & 34.4 & \textbf{2.1} & 21.9 & 61.5 & 23.9 & 32.3 & 35.3 & 4.7 & \textbf{15.2} & \textbf{7.7} & \textbf{8.5} & 24.7\\
\cline{4-17}
 & Pred &  & 32.2 & 33.5 & 32.0 & \textbf{2.1} & 16.1 & 62.0 & 23.5 & 28.1 & 30.4 & \textbf{6.2} & 13.6 & 7.6 & 6.2 & 22.6\\
\hline
\hline
\multirow{3}{*}{Intensity} & Grid & \multirow{3}{*}{GNN} & 43.4 & 38.8 & 36.0 & 3.7 & 21.3 & 71.4 & 25.7 & 37.9 & 32.6 & 6.5 & 12.4 & 9.2 & 7.1 & 26.6\\
\cline{4-17}
 & Key-points &  & 44.1 & 36.9 & 38.8 & 3.3 & 19.1 & 66.0 & 26.5 & 32.9 & 33.3 & 5.9 & \textbf{13.7} & 7.2 & 7.5 & 25.8\\
\cline{4-17}
 & Pred &  & \textbf{50.1} & \textbf{49.4} & \textbf{49.5} & \textbf{5.4} & \textbf{22.9} & \textbf{74.6} & \textbf{33.2} & \textbf{38.3} & \textbf{38.5} & \textbf{10.0} & \textbf{13.7} & \textbf{13.6} & \textbf{9.3} & \textbf{31.4}\\
\hline
\hline
\multirow{3}{*}{MIND} & Grid & \multirow{3}{*}{Mean Field} & \textbf{48.9} & \textbf{47.0} & \textbf{47.5} & 1.2 & 13.1 & \textbf{68.8} & \textbf{28.4} & 29.0 & 25.6 & 4.6 & 1.0 & 7.4 & 6.7 & 26.0\\
\cline{4-17}
 & Key-points &  & 40.9 & 33.9 & 35.0 & 2.1 & \textbf{21.0} & 61.2 & 23.9 & \textbf{32.6} & \textbf{34.8} & 4.6 & \textbf{15.0} & 7.5 & \textbf{8.7} & 24.7\\
\cline{4-17}
 & Pred &  & 44.8 & 45.4 & 44.9 & \textbf{2.1} & 15.4 & 67.0 & \textbf{28.4} & 29.6 & 31.2 & \textbf{5.3} & 12.7 & \textbf{10.0} & 8.5 & \textbf{26.6}\\
\hline
\hline
\multirow{3}{*}{MIND} & Grid & \multirow{3}{*}{GNN} & 47.4 & 46.2 & 45.1 & 3.3 & 17.7 & 67.6 & 30.9 & 36.6 & 30.6 & 7.2 & 13.7 & 8.3 & 8.3 & 27.9\\
\cline{4-17}
 & Key-points &  & 48.0 & 43.4 & 44.0 & 1.9 & 18.8 & 66.0 & 28.8 & 35.0 & 33.1 & 7.1 & 14.7 & 6.6 & 8.1 & 27.3\\
\cline{4-17}
 & Pred &  & \textbf{53.5} & \textbf{46.4} & \textbf{53.4} & \textbf{3.9} & \textbf{23.9} & \textbf{72.1} & \textbf{33.6} & \textbf{40.4} & \textbf{41.1} & \textbf{10.1} & \textbf{17.3} & \textbf{10.4} & \textbf{10.3} & \textbf{32.0}\\
\hline
\hline
\multirow{3}{*}{UNet} & Grid & \multirow{3}{*}{Mean-Field} & 48.4 & 38.8 & 37.0 & 2.4 & 21.2 & 69.0 & 28.1 & 38.3 & 39.0 & 7.3 & 16.7 & 9.6 & \textbf{11.6} & 28.3\\
\cline{4-17}
 & Key-points &  & \textbf{53.0} & 42.2 & \textbf{42.0} & 2.6 & 25.5 & 69.9 & \textbf{33.0} & 37.8 & 38.7 & 8.9 & 17.1 & 11.3 & 11.3 & 30.3\\
\cline{4-17}
 & Pred &  & 51.5 & \textbf{47.4} & 40.1 & \textbf{3.7} & \textbf{29.2} & \textbf{74.0} & 31.3 & \textbf{40.7} & \textbf{42.6} & \textbf{9.4} & \textbf{18.8} & \textbf{13.2} & 11.2 & \textbf{31.8}\\
\hline
\hline
\multirow{3}{*}{UNet} & Grid & \multirow{3}{*}{GNN} & 57.8 & 51.6 & 54.8 & 2.8 & 23.2 & 73.8 & 33.6 & 44.8 & 47.6 & 11.0 & 18.2 & 17.6 & 11.6 & 34.5\\
\cline{4-17}
 & Key-points & & 57.9 & 52.5 & 52.0 & 3.7 & \textbf{27.5} & 75.4 & 38.4 & 44.7 & 47.3 & 12.2 & 18.7 & 16.2 & 9.8 & 35.1\\
\cline{4-17}
 & Pred & & \textbf{62.9} & \textbf{62.5} & \textbf{59.8} & \textbf{4.6} & 27.4 & \textbf{81.6} & \textbf{37.0} & \textbf{51.1} & \textbf{49.7} & \textbf{13.5} & \textbf{21.3} & \textbf{20.8} & \textbf{14.5} & \textbf{39.0} \\
\hline

\end{tabular}
\caption{Dice score per structure for the CT dataset. The considered structures are: the spleen~\crule[green]{0.3cm}{0.3cm} , the right kidney~\crule[red]{0.3cm}{0.3cm}, the left kidney~\crule[blue]{0.3cm}{0.3cm}, the gallblader~\crule[yellow]{0.3cm}{0.3cm}, the esophagus~\crule[purple]{0.3cm}{0.3cm}, the liver~\crule[pink]{0.3cm}{0.3cm}, the stomach~\crule[orange]{0.3cm}{0.3cm}, the aorta~\crule[brown]{0.3cm}{0.3cm}, the inferior vena cava~\crule[cyan]{0.3cm}{0.3cm}, the portal and splenic veines~\crule[magenta]{0.3cm}{0.3cm}, the pancreas~\crule[olive]{0.3cm}{0.3cm}, the right adrenal gland~\crule[teal]{0.3cm}{0.3cm}, the left adrenal gland~\crule[gray]{0.3cm}{0.3cm}.}
\label{Exp1}
\end{table}

\end{landscape}

\begin{table}[H]
    \centering
    \begin{tabular}{|c|c|c||c|c|c|c|c|}
        \hline
         \multicolumn{3}{|c|}{Configuration choice} &  \multicolumn{5}{|c|}{Dice}\\
         \hline
         \rotatebox[origin=c]{70}{Features} & \rotatebox[origin=c]{70}{Driving points} & \rotatebox[origin=c]{70}{Regularization} &  \crule[green]{0.3cm}{0.3cm} & \crule[red]{0.3cm}{0.3cm} & \crule[blue]{0.3cm}{0.3cm} &  \crule[pink]{0.3cm}{0.3cm} & mean\\
         \hline
         \hline
\multirow{3}{*}{Intensity} & Grid & \multirow{3}{*}{Mean Field}  & 26.4 & 40.8 & \textbf{35.1} & 65.1 & 41.8\\
\cline{4-8}
 & Key-points &  & 26.1 & 40.9 & 34.6 & 65.1 & 41.7\\
\cline{4-8}
 & Pred &  & \textbf{29.3} & \textbf{48.7} & 35.0 & \textbf{66.5} & \textbf{44.9}\\
\hline

\multirow{3}{*}{Intensity} & Grid & \multirow{3}{*}{GNN} & 44.4 & 64.6 & 51.7 & 74.5 & 58.8\\
\cline{4-8}
 & Key-points &  & 50.1 & 63.1 & 49.0 & 72.7 & 58.7\\
\cline{4-8}
 & Pred &  & \textbf{56.0} & \textbf{70.5} & \textbf{58.7} & \textbf{77.6} & \textbf{65.7}\\
\hline

\multirow{3}{*}{MIND} & Grid & \multirow{3}{*}{Mean Field} & 26.5 & 40.4 & 35.2 & 65.0 & 41.8\\
\cline{4-8}
 & Key-points &  & 26.5 & 40.6 & 35.0 & 65.0 & 41.8\\
\cline{4-8}
 & Pred &  & \textbf{45.3} & \textbf{61.1} & \textbf{51.3} & \textbf{73.9} & \textbf{57.9}\\
\hline

\multirow{3}{*}{MIND} & Grid & \multirow{3}{*}{GNN} & 47.3 & 68.3 & 56.0 & 76.6 & 62.1\\
\cline{4-8}
 & Key-points &  & 50.3 & 65.1 & 54.4 & 73.8 & 60.9\\
\cline{4-8}
 & Pred &  & \textbf{56.6} & \textbf{73.4} & \textbf{63.4} & \textbf{78.4} & \textbf{67.9}\\
\hline

\multirow{3}{*}{UNet} & Grid & \multirow{3}{*}{Mean Field} & 61.1 & 70.5 & 63.0 & 79.3 & 68.5\\
\cline{4-8}
 & Key-points &  & 64.1 & 73.1 & 65.2 & 77.7 & 70.0\\
\cline{4-8}
 & Pred &  & \textbf{64.4} & \textbf{74.3} & \textbf{66.3} & \textbf{80.6} & \textbf{71.4}\\
\hline

\multirow{3}{*}{UNet} & Grid & \multirow{3}{*}{GNN} & 63.8 & 77.0 & 63.0 & 78.2 & 72.5\\
\cline{4-8}
 & Key-points &  & 65.4 & 76.1 & 67.9 & 81.0 & 72.6\\
\cline{4-8}
 & Pred &  & \textbf{72.0} & \textbf{80.3} & \textbf{72.4} & \textbf{83.4} & \textbf{77.0}\\
\hline
\end{tabular}
\caption{Dice score per structure for the MR dataset. The considered structures are: the spleen~\crule[green]{0.3cm}{0.3cm} , the right kidney~\crule[red]{0.3cm}{0.3cm}, the left kidney~\crule[blue]{0.3cm}{0.3cm} and the liver~\crule[pink]{0.3cm}{0.3cm}}
\label{Exp2}
\end{table}

\begin{figure}
\centering
     \begin{subfigure}[b]{0.48\linewidth}
         \centering
         \includegraphics[width=\linewidth]{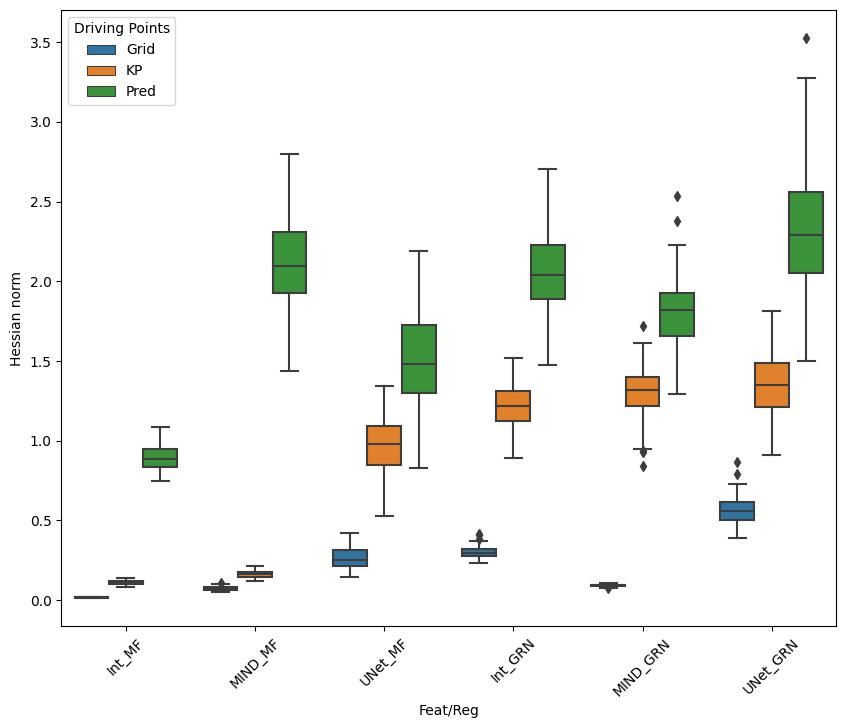}
         \caption{CT dataset}
         \label{Hessian L2R}
     \end{subfigure}
    \centering
     \begin{subfigure}[b]{0.48\linewidth}
         \centering
         \includegraphics[width=\linewidth]{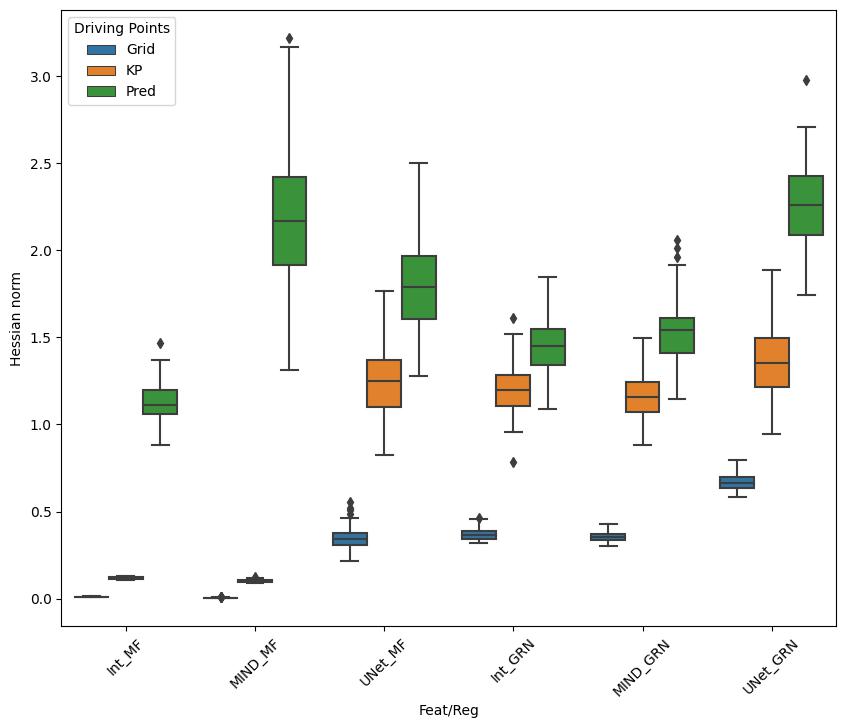}
         \caption{MR dataset}
         \label{Hessian CHAOS}
     \end{subfigure}
     \hfill
\caption{Mean Hessian Norm for the predicted displacement fields}
\label{Hessian}
\end{figure}

\begin{figure}
\centering
     \begin{subfigure}[b]{0.48\linewidth}
         \centering
         \includegraphics[width=\linewidth]{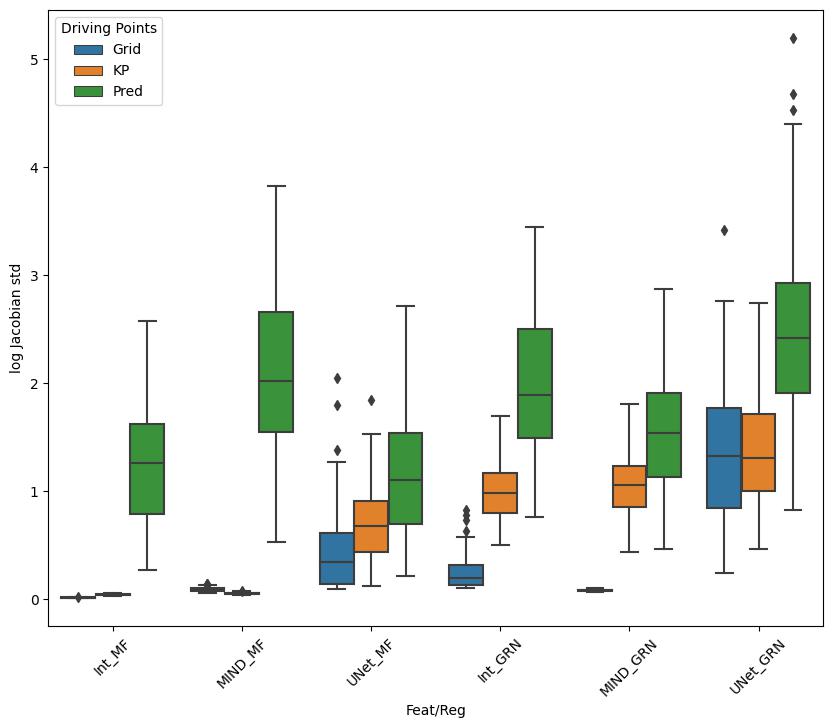}
         \caption{CT dataset}
         \label{STDlogJ L2R}
     \end{subfigure}
    \centering
     \begin{subfigure}[b]{0.48\linewidth}
         \centering
         \includegraphics[width=\linewidth]{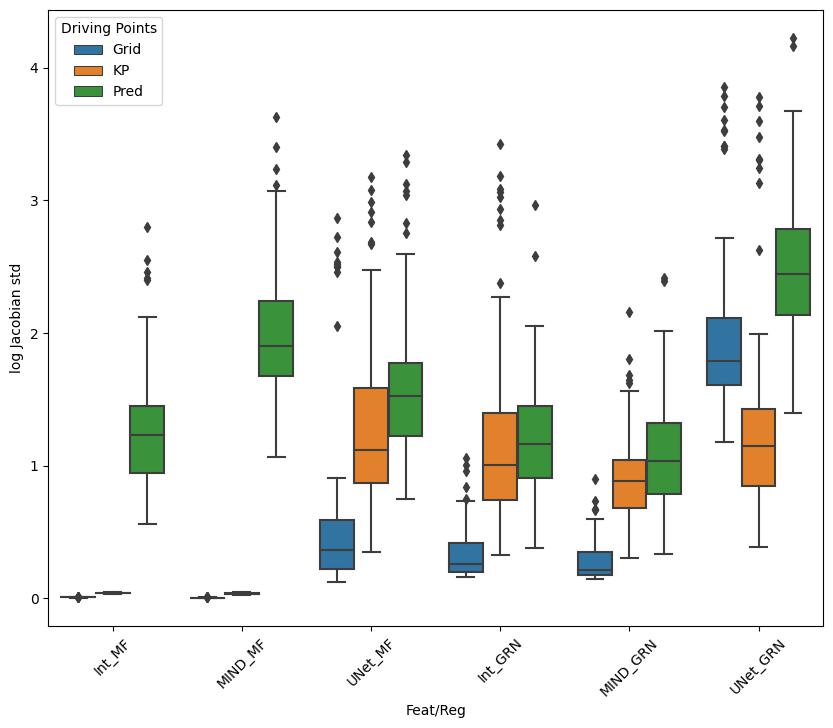}
         \caption{MR dataset}
         \label{STDlogJ CHAOS}
     \end{subfigure}
     \hfill
\caption{Standard deviation of the log Jacobian determinant of the predicted deformation fields}
\label{STDlogJ}
\end{figure}

\begin{figure}
\centering
 \includegraphics[width=\linewidth]{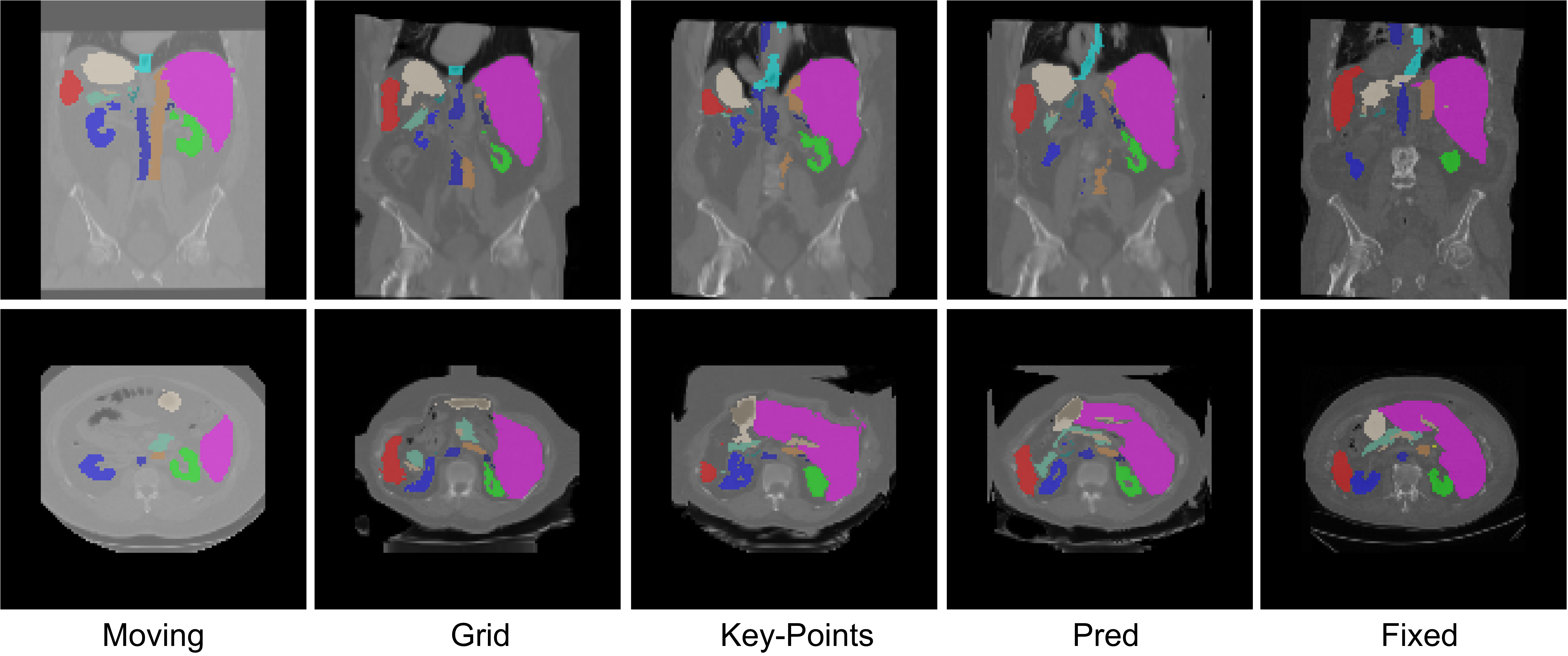}
 \caption{Visual comparison of registered images from the CT dataset for different driving points selection methods using a deep feature extractor and GRN regularization.}
 \label{Exp1_viz}
\end{figure}

\begin{figure}
\centering
 \includegraphics[width=\linewidth]{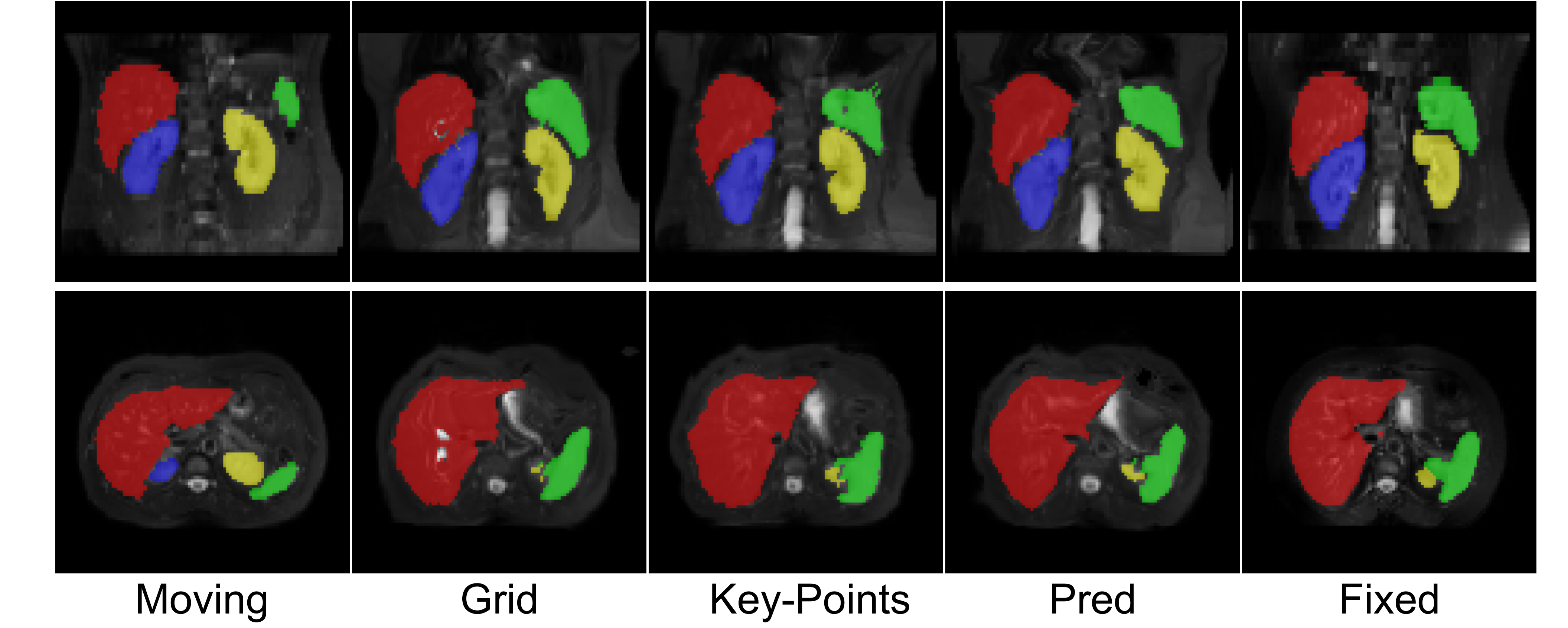}
 \caption{Visual comparison of registered images from the MR dataset for different driving points selection methods using a deep feature extractor and GRN regularization.}
 \label{Exp2_viz}
\end{figure}

\begin{figure}
\centering
 \includegraphics[width=\linewidth]{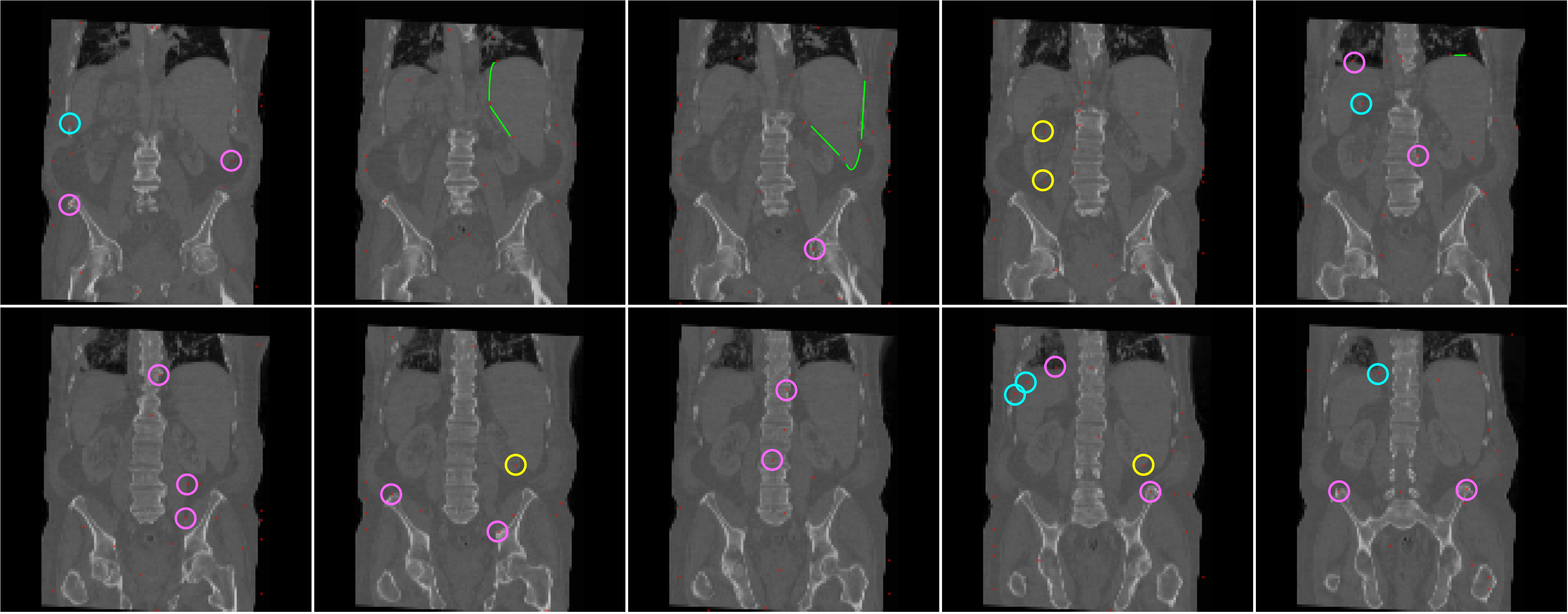}
 \caption{Predicted driving points (small red dots) predicted for a pair of images to be registered from the CT dataset. 10 consecutive coronal slices (left-to-right then top-to-bottom) of the fixed patient are shown. The pink circles highlight driving points that could be considered as key-points as they are salient points (tips of bones or organs). In particular, the yellow points show that the tips of both kidneys are selected as driving points. Similarly, we connected with green segments predicted driving points that all together parameterize the deformation of the liver border. The predicted driving points highlighted in cyan form a contour of the spleen. We also note that areas with poor contrast are less densely populated with driving points.}
 \label{Driving points illustrated}
\end{figure}

\begin{figure}
\centering
     \begin{subfigure}[b]{0.48\linewidth}
         \centering
         \includegraphics[width=\linewidth]{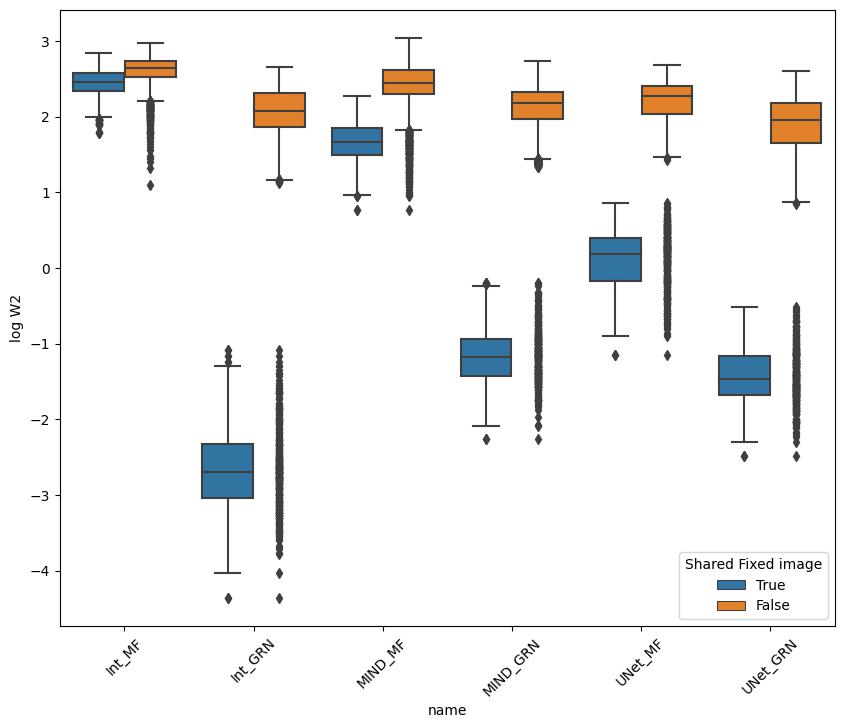}
         \caption{CT dataset}
         \label{Mean W2 L2R}
     \end{subfigure}
    \centering
     \begin{subfigure}[b]{0.48\linewidth}
         \centering
         \includegraphics[width=\linewidth]{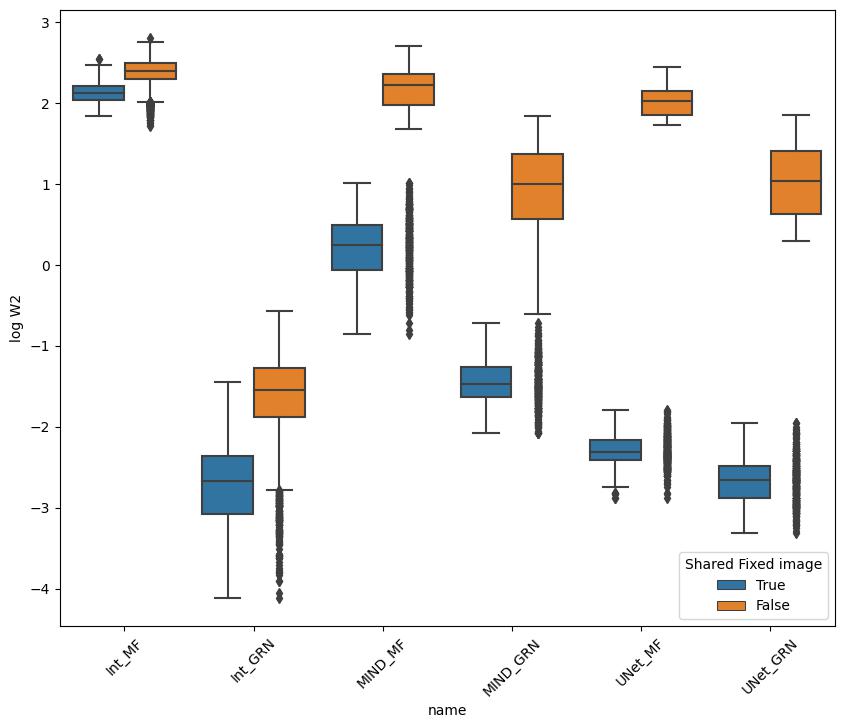}
         \caption{MR dataset}
         \label{Mean W2 CHAOS}
     \end{subfigure}
     \hfill
\caption{Log W2 distance between predicted driving point sets for two pairs of images to be registered sharing the same fixed image (Shared Fixed image is True) versus Log w2 distance between all predicted driving point sets (Shared Fixed image is False). We observe that the mean W2 distance between driving point sets predicted for inputs sharing the same fixed image is systematically lower than the mean W2 distance between all predicted driving point sets. We also observe that the mean W2 distance between predicted driving point sets for inputs sharing the same fixed image is smaller but non-zero which shows that the obtained driving points predictors marginally adapt their prediction to the moving image data as well.}
\label{Mean W2}
\end{figure}

\end{document}